\documentclass[letterpaper, 10 pt, conference]{ieeeconf}
\IEEEoverridecommandlockouts
\usepackage{graphicx} 
\usepackage{amsmath}
\usepackage{amsfonts}
\usepackage{amssymb}
\usepackage{xcolor}
\usepackage{subcaption}
\usepackage{algpseudocode}
\usepackage{algorithm}
\usepackage{cite}
\usepackage[hyphens]{url}
\usepackage[colorlinks=true, allcolors=blue]{hyperref}

\newcommand{\PH}[1]{\noindent\hphantom{\textbf{#1:} }\ignorespaces}

\newcommand{\Var}{\text{Var}}

\newcommand{\figref}[1]{Fig. \ref{#1}}

\title{\LARGE \bf{Simulated Annealing for Multi-Robot Ergodic Information Acquisition Using Graph-Based Discretization}
}

\author{Benjamin Wong$^{1}$, Aaron Weber$^{1}$, Mohamed M. Safwat$^{1}$, Santosh Devasia$^{1}$, and Ashis G. Banerjee$^{2}$
\thanks{$^{1}$B. Wong, A. Weber, M. M. Safwat, and S. Devasia are with the Department of Mechanical Engineering, University of Washington, Seattle, WA 98195, USA.
{\tt\small bycw,aweber6,mohsaf,devasia@uw.edu}}%
\thanks{$^{2}$A. G. Banerjee is with the Department of Industrial \& Systems Engineering and Department of Mechanical Engineering, University of Washington, Seattle, WA 98195, USA.
{\tt\small ashisb@uw.edu}}%
}

\begin{document}

\maketitle

\begin{abstract}

One of the goals of active information acquisition using multi-robot teams is 
to keep the relative uncertainty in each region at the same level to maintain identical acquisition quality (e.g., consistent target detection) in all the regions. To achieve this goal, ergodic coverage can be used to assign the number of samples according to the quality of observation, i.e., sampling noise levels. However, the noise levels are unknown to the robots. Although this noise can be estimated from samples, the estimates are unreliable at first and can generate fluctuating values. The main contribution of this paper is to use simulated annealing to generate the target sampling distribution, 
starting from uniform and gradually shifting to an estimated optimal distribution, 
by varying the coldness parameter of a Boltzmann distribution with the estimated sampling entropy as energy. Simulation results show a substantial improvement of both transient and asymptotic entropy compared to both uniform and direct-ergodic searches. Finally, a demonstration is performed with a TurtleBot swarm system to validate the physical applicability of the algorithm. 

\end{abstract}

\section{Introduction}

The robotics community has been interested in multi-robot systems for their versatility and efficiency in performing time consuming and repetitive tasks in a parallel manner. Many of these tasks belong to active information acquisition, including surveillance, inspection, environmental monitoring, and disaster response \cite{9627691, WongBenjamin2023Hrdo,9812651, 8866581}. In these scenarios, a team of robots is tasked with collecting or gathering information by traveling between various sites, often to locate targets, such as defects in inspection operations, survivors in disaster responses, or sources of hazard during surveillance and environmental monitoring. By using a multi-robot system, resources can be distributed (allocated) among different regions to effectively perform time-critical tasks.

\textit{Ergodic control} is well suited for resource allocation in such acquisition tasks, which amount to sampling from a statistical perspective.
This is particularly true for a cost-effective multi-robot system, where a large team of robots are
equipped with commodity sensors that provide uncertain measurements.
These uncertain (noisy) measurements are expressed in terms of probability, where estimates can be improved by repeated measurements. Moreover, the confidence of the estimates is directly correlated to the measurement noise (e.g., the target is hiding in a visually cluttered environment as opposed to a target in plain sight), such that more samples are needed for a noisier target to reach the same level of confidence as a less noisy target. This provides an ideal number of total visitation to each target site (region) that the team of robots has to maintain as a whole. This aligns with the core objective of ergodic control,
where a control law is devised such that the time-averaged visitation frequency is equal to the (spatial) target distribution \cite{MathewGeorge2011Mfea}, which is specified by information quality in the case of sampling.

The main challenge for applying ergodic control though 
is that the information quality is initially unknown to the robot team. In traditional ergodic control, the target distribution is assumed to be known, either from an oracle, prior experience, or experts' demonstrations. In the case of optimal sampling, the sampling variances are initially unknown to the robots. Hence, the optimal target distribution cannot be derived. The sample variance can be estimated through the collected samples as the robots start exploring. However, it is unreliable when the sample size is small and fluctuates 
as new samples are incorporated. This leads to a highly time-varying and unreliable target distribution that causes further misallocation of resources. 

To avoid this problem, the target distribution should focus on collecting information on the variance at the beginning and gradually shift toward the optimal solution as the estimated variance becomes reliable. This is done by applying \textit{simulated annealing} to the target distribution. Annealing is a process of heating up a material and slowly cooling it down to allow the molecules to reach the lowest energy state \cite{HumphreysF.J.2017Rara}. Simulated annealing applies this idea to stochastic optimization to encourage exploration by introducing substantial randomness (high entropy) at the beginning and gradually reducing the randomness until the optimal solution is reached. In this work, the robot team is treated as a particle swarm with the noise level of each region corresponding to the energy state. The high entropy state corresponds to the robots spreading out to all the regions regardless of the estimated noise level, and low entropy corresponds to all the robots focusing on the highest noise level region. This allows for a smooth interpolation between a uniform target distribution and a greedy distribution. Accordingly, the robot team is able to collect information about the variance and smoothly transition to the optimal distribution. The main contributions of this work are summarized below.

\begin{itemize}
    \item Formulating a multi-robot information acquisition problem where ergodic control provides an optimal solution
    \item Incorporating simulated annealing to handle unreliable information (noisy measurements) in the ergodic controller 
    \item Demonstrating superior performance in terms of posterior entropy compared to direct and uniform ergodic methods 
\end{itemize}

\section{Related Works}
\subsection{Active Sampling}
In a broad sense, this work falls in the domain of active planning, where agents plan trajectories that satisfy an objective and continuously adjust their trajectories as new information arrives. In this case, the objective is to obtain information through sampling. A popular choice of doing so is to find a trajectory that covers the entire space \cite{9231484, 8256741, 1545323, ShahNikunj2025EFfR}. This choice assumes that all the points in the space are equally important and lacks the flexibility to adapt to the quality of information. Alternatively, to consider information density, sensor placement through Voronoi partitioning has been used \cite{CortesJ.2004Ccfm, JingLi2009VBAC, DoodmanSaeid2025AnVo, MannaPrithwish2025ACFf}. This is well-suited for a static placement problem, where the robots converge to the optimal position until a new event arises. This requires a known number of robots and assumes all robots to be active, 
and does not account for the need of redundancy. On the other hand, in ergodic control, all the agents provide coverage individually, which make the system more robust to sensor and actuation failures and varying numbers of active robots.   

Our formulation of obtaining information from discrete locations is essentially a problem of sequential experiment design \cite{RobbinsHerbert1952Saot}. A particular form that has been thoroughly studied is the multi-armed bandit (MAB) problem, where the agents maximize their rewards under unknown distributions \cite{SlivkinsAleksandrs2024ItMB}. Traditionally, MAB assumes that any arm can be sampled at any time. Recently, work has been done on applying the MAB problem constrained on a graph \cite{ZhangTianpeng2023MBLo}, similar to our problem formulation. 
While we can consider the information as a form of reward in our formulation, the biggest difference with the MAB problem is that the MAB rewards are fungible, where agents can collect solely from the state with the maximum amount of reward. On the other hand, in our formulation, 
sufficient information has to be collected from all the regions, i.e. missing information in one region cannot be substituted by more information from another region.

\subsection{Ergodic Control}
In ergodic control, a controller is designed such that the time-averaged trajectories of the agents equal spatial target distributions. It is, therefore, often used in robotics to solve multi-agent resource allocation and coverage problems \cite{IvicStefan2017ECMA, SalmanHadi2017MECw, IvicStefan2020Ssia, PatelShivang2021MECi, IvicStefan2022Cmea, LerchCameron2023SEEi, XuAlbert2024MPFf}. Early foundational works assumed the target distribution to be given \cite{MathewGeorge2011Mfea, AyvaliElif2017Ecic, IvicStefan2022Cmea}. Other application-oriented works considered ergodic control as a means to promote exploration by assigning a distribution around the optimal \cite{MillerLaurenM2016EEoD, AbrahamIan2021AEMf, PignatEmmanuel2022Lfdu, ShettySuhan2022EEUT}. Recently, discrete ergodic control formulations have been investigated for robot exploration in topologically challenging environments \cite{CrnkovicBojan2023Fafc,ShiroseBurhanuddin2024GGES, WongBenjamin2025RCTE}. In this work, we aim to extend such graph-based formulations to address a specific multi-robot allocation problem, where the 
target distribution is modeled as an objective for the ergodic controller. In other words, our ergodic controller is designed to yield an optimal strategy for allocating robots to different regions so as to acquire information in a Bayesian manner.

\subsection{Simulated Annealing}
This work utilizes simulated annealing to transition from exploration to exploitation. Simulated annealing originated from statistical mechanics and is widely applied to complex optimization problems such as the traveling salesman problem \cite{KirkpatrickS.1983ObSA}. It is rooted in the study of ergodic systems, with the most common implementation being the Metropolis-Hasting (M-H) algorithm as a stochastic alternative to gradient descent \cite[chapter 8]{BremaudPierre2013MCGF}. It has been applied to robotic systems for path planning problems as a means to escape local minima \cite{4913141, 9750552, 8793838, 11127661, QidanZhu2006RPPB}. In these works, the target distribution is implicitly defined under the acceptance-rejection step of M-H. In comparison, as an ergodic control framework, the (annealed) target distribution in our work is explicitly defined, which yields a method that has more desired characteristics, such as convergence rate.

\section{Problem Statement}
\subsection{Problem Formulation}
    \begin{figure}[thpb]
      \centering
       \framebox{\parbox{0.3\textwidth}{
       \includegraphics[width=0.3\textwidth]{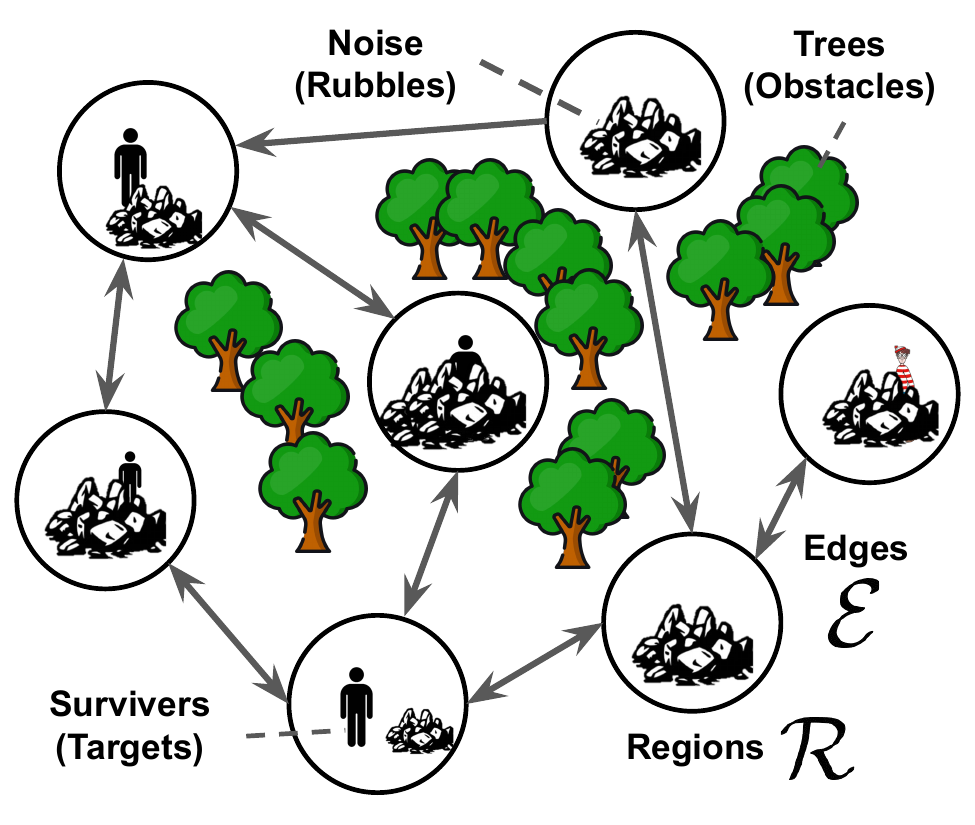}
       }}
      \caption{Example information gathering task of locating survivors in a map \(\mathcal{G}\) with regions \(\mathcal{R}\) and edges \(\mathcal{E}\) caused by blockage of trees. The regions containing various amount of rubble, which causes differences in information quality i.e., noise levels \(\sigma_i^2\), across the regions.
      }
      \label{fig:example_task}
   \end{figure}
    This article considers information gathering tasks that are scattered in a large area, with regions of interest separated by obstacles. A team of \(N\) robots has to estimate the states internal to each region. An example task of locating survivors in disaster relief is shown in \figref{fig:example_task}, where the states to be estimated are the survivors' locations in all the regions.
    
    The regions are modeled as a set of nodes \(\mathcal{R}\).  Connections between pairs of regions are represented by an edge, and \(\mathcal{E}\) is the set of all the edges. Then, the tuple of nodes and edges forms the graph \(\mathcal{G} = (\mathcal{R}, \mathcal{E})\). 
     The information gathering is considered to be estimating the scalar state \(x_i\) for each region \(r_i \in \mathcal{R}\). Each robot can collect observation from the region at every time step, with the noisy observation model
    \begin{equation}
    \label{eq:observation}
    \begin{aligned}
    z_i = x_i + \epsilon_i, \quad \epsilon_i \sim\mathcal{N}(0, \sigma_i^2).
    \end{aligned}
    \end{equation}
    The variance \(\sigma_i^2\) varies between the regions but not between the robots, i.e., some regions have lower information quality per observation than others. The varying variance can be an effect from the lack of landmark for the robot to localize or high visual noise in the region that conceals the target. The state \(x_i\) can be estimated from the observations by  
    \begin{equation}
    \label{eq:mean_estimate}
    \begin{aligned}
    \bar{x}_i = \frac{1}{\nu_i}\sum^{\nu_i}_{j=1} z_j \approx x_i, \quad \text{Var}[\bar{x}_i ]= \frac{\sigma^2_i}{\nu_i}
    \end{aligned}
    \end{equation}
    where \(\nu_i\) is the number of observations acquired by the robots collectively. With the standard error, i.e. the variance of the estimation \(\Var(\bar{x_i})\),
    being a function of the number of observation \(\nu_i\) and the regional noise \(\sigma_i^2\) \footnote{By the central limit theorem, this is also extendable to non-Gaussian distribution with a sufficiently large \(\nu\).}

\subsection{Optimal Distribution}
    
    To ensure no one region has more standard error than the others, the goal is to minimize the maximum posterior variance after any time horizon \(K\), i.e. all regions are equally confident at all time,
    \begin{equation}
    \begin{aligned}
    \label{eq:optimal_dist_problem}
    \min_{\bar\rho} &\quad \max_i \left [\frac{\sigma^2_i}{KN\bar\rho_i} \right]\\
    \text{s.t.} &\quad \mathbf{1}^T\bar\rho = 1  \\
    &\quad  \bar\rho  \succeq 0
    \end{aligned}
    \end{equation}
    where \(\bar\rho_i\) is the relative visitation frequency of region \(i\), and \(\bar\rho\) is the target distribution, which is a vector with the \(i\)-th entry being \(\bar\rho_i\) ; \(N\) is the number of robots. For a known sample variance \(\sigma^2\), the optimal target distribution is 
    \begin{equation}
    \label{eq:optimal_dist}
    \begin{aligned}
    \bar{\rho}^*_i = \frac{1}{\sum_i \sigma^2_i}\sigma^2_i.
    \end{aligned}
    \end{equation} 
    Then, by substituting \(\bar\rho^*\) to \(\bar\rho\) in \eqref{eq:optimal_dist_problem}, the variance of the estimation at any time step \(K\) is\footnote{It can be observed that uniform values minimize the maximum value since to maintain a constant sum, any decrease of value in one region will cause an increase of equal amount in other regions, which increases the maximum value. }
    \begin{equation}
    \begin{aligned}
    \Var[\bar{x}_i] =  \frac{\sigma^2_i}{KN\bar\rho_i} = \frac{\sum_i\sigma^2_i}{KN}
    \end{aligned}
    \end{equation}
    which is the same for all regions. Hence, there exists a need for a planning strategy such that the target \(\bar\rho\) can be reached under the constraint of the graph connectivity.

    Moreover, the true sampling variance \(\sigma^2_i\) are initially unknown to all the agents. The variance can also be estimated along side \(x_i\) as the robots start collecting samples, which is done by
    \begin{equation}
    \begin{aligned}
    \label{eq:variance_estimate}
    \bar\sigma_i^2 = \frac{1}{\nu_i-1}\sum_{j=1}^{\nu_i} (z_j - x_i)^2 \approx \sigma_i^2.
    \end{aligned}
    \end{equation}
    Similar to the mean \(\bar{x}\), 
    the estimated variance \(\bar\sigma\)
    is unreliable at the beginning and are improved with increasing number of samples. As a result, $\bar{\rho}^*$ in \eqref{eq:optimal_dist} is unknown to the robots and the \(\bar{\rho}\) generated by the estimated variance \(\bar\sigma^2\) is unreliable and causes inefficient allocation. This motivates the research problem to develop
    a planning method that can utilize the estimated variance while 
    accounting for
    the unreliable initial estimations.
   This can be considered as two subproblems in developing  the planning method: (subproblem 1)~to reach a given target distribution on a graph with a multi-robot system and (subproblem 2)~to account for the unknown variance and unreliable estimates in the beginning.
    
\section{Methodology}
 \begin{figure}[thpb]
      \centering
       \framebox{\parbox{0.45\textwidth}{
       \includegraphics[width=0.45\textwidth]{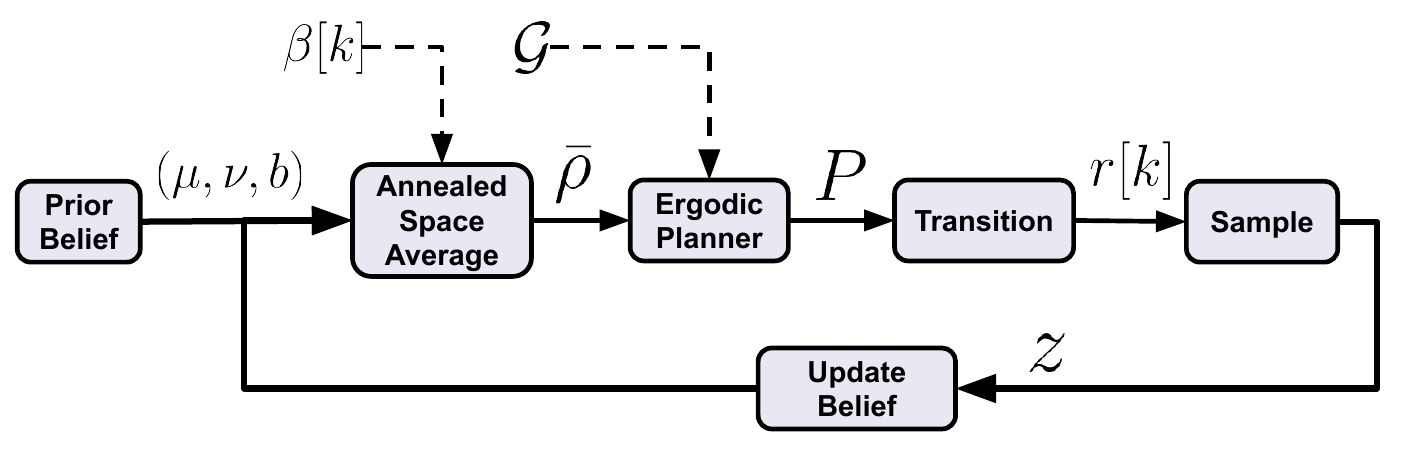}
       }}
      \caption{Flowchart of the annealed ergodic information gathering algorithm.}
      \label{fig:flowchart}
   \end{figure}

The problem of achieving the target distribution is solved by using \textit{Rapidly Ergodic Markov Chain} (REMC), and the problem of unreliable initial estimated variance is solved by using \textit{simulated annealing}. The solution methods are explained in Sections \ref{section:MREC} and \ref{section:annealing}, respectively. The overall pipeline of the planning method is shown in \figref{fig:flowchart}. 

\subsection{Subproblem 1 Solution: Multi-Robot Ergodic Control }
\label{section:MREC}

    Ergodic control can solve the subproblem 1 to
    achieve the target distribution \(\bar\rho\) for a given \(\sigma^2\). In general, the goal for ergodic control is to synthesize a control law such that the dynamic system has a time average equal to the space average for almost all initial conditions. This is formulated in a graph space as
    \begin{equation}
    \label{eq:graph_ergodicity}
         \lim_{K \to \infty} \frac{1}{K}  \sum_{k=0}^{K-1} \left(F(r[k])\right )   = \frac{1}{\mu(\mathcal{R})}  \sum_{r_i \in \mathcal{R}}F(r_i)\mu(r_i) 
    \end{equation} 
    where \(r[\cdot]\) is the region trajectory; \(\mathcal{R}\) is the set of regions; \(\mu\) is a measure on the region set; and \(F\) is an arbitrary \(\mu\)-measurable function. To measure visitation, \(F\) can be defined as the indicator function 
    \begin{equation}
    \label{indicator_function}
    \begin{aligned}
     I(r_i) &\triangleq \begin{bmatrix}
     \delta_{1,i} &
     \delta_{2,i} &
     \cdots&
     \delta_{n,i}
    \end{bmatrix}^T
    \end{aligned}
    \end{equation} 
    where \(\delta_{i,j}\) is the 
    Kronecker delta, and
    \begin{equation}
    \label{eq:space_time_average}
    \begin{aligned}
         \hat{\rho}   \triangleq \lim_{K \to \infty} \frac{1}{K}  \sum_{k=0}^{K-1} \left(I(r[k])\right ), \quad   \bar{\rho}  \triangleq \frac{1}{\mu(\mathcal{R})} \sum_{r_i \in \mathcal{R}}I(r_i)\mu(r_i).
    \end{aligned}
    \end{equation} 
    The ergodic objective for a single agent has been shown to be achieved by generating the trajectory with \textit{Markov chains} \cite{WongBenjamin2025RCTE}. The Markov chain is generated by randomly sampling the next region based only on the current region, i.e. \(r[k+1] \sim \mathbb{P}(R\ |\ r[k])\), with \(R\) being the random variable of the possible region. In a finite graph space, the transition probability can be represented by a stochastic matrix \(P\). The time average can then be expressed in terms of expected value as
    \begin{equation}
    \begin{aligned}
    \label{eq:time_average_markov}
         \mathbb{E}\left[\hat\rho\right] &= \lim_{K \to \infty} \frac{1}{K}  \sum_{k=0}^{K-1} \left(I(\mathbb{E}[r[k]])\right )  \\
         &=\lim_{K \to \infty} \frac{1}{K}  \sum_{k=0}^{K-1} \left(P^k\rho[0]\right ) 
    \end{aligned}
    \end{equation} 
    with \(\rho[0]\) as the initial distribution. 
    
    The transition matrix that guarantees ergodicity and also optimizes the convergence rate can be found by the following convex program:
    \begin{equation}
    \label{eq:remc}
    \begin{aligned}
        \arg\min_P & \quad \lambda_{\text{max}}\left(\frac{1}{2}\left(\tilde{P}+\tilde{P}^T\right)-2\bar{\rho}^{1/2}\bar{\rho}^{T/2}\right) 
    \\
         \text{s.t.} &\quad  \textbf{1}^TP = \textbf{1}^T \quad (\text{Stochastic }P)\\
            & \quad P\bar{\rho} = \bar{\rho} \quad (\text{Target Distribution})\\
            & \quad P_{i,j} \geq 0 \quad (\text{Stochastic }P)\\
            & \quad P_{i,j} = 0 \quad \text{if} \quad (j,i) \notin \mathcal{E} \quad {(\mbox{Transitions in}}~\mathcal{G}) \\
            & \quad \tilde{P} = \text{diag}(\bar\rho^{-1/2})P\text{diag}(\bar\rho^{1/2}).
    \end{aligned}
    \end{equation}
    
    As ergodicity is a property derived from ensemble systems, \eqref{eq:graph_ergodicity} is extensible to multi-robot system. When all the robots follow the same Markov chain, \eqref{eq:space_time_average} is modified to account for the trajectories of all the robots as 
    \begin{equation}
    \begin{aligned}
        \mathbb{E}[\hat\rho] & =  \frac{1}{N}\sum_{a=1}^N \mathbb{E}[\hat{\rho}_a]\\
        & =\frac{1}{N}\sum_{a=1}^N\left(\lim_{K\to\infty} \frac{1}{K}\sum_{k=0}^{K-1} P^k\rho_a[0] \right) \\
        & = \lim_{K\to\infty} \frac{1}{K}\sum_{k=0}^{K-1}\left( P^k\frac{1}{N}\sum_{a=1}^N\rho_a[0] \right) \\
        & =  \lim_{K\to\infty} \frac{1}{K}\sum_{k=0}^{K-1}\left( P^k\rho[0]\right). 
    \end{aligned}
    \end{equation}
    Here, \(\rho_a\) denotes the distribution for agent \(a\), and \(\rho\) is averaged over all the robots. Consequently, the more robots are in the team, the less variance the Markov chain would have. That is, the true trajectory follows more closely to the expected value
        \begin{equation}
            \frac{1}{N}\sum_{a=1}^{N} I(r_a[k]) \rightarrow{} P^k\rho[0] \quad \text{as} \quad N\rightarrow\infty.
        \end{equation}

\subsection{Subproblem 2 Solution: Annealing for Space Average }
\label{section:annealing}
    \begin{figure*}[thpb!]
        \centering
         \begin{subfigure}[t]{0.22\textwidth}
            \centering
            \includegraphics[width=1\textwidth]{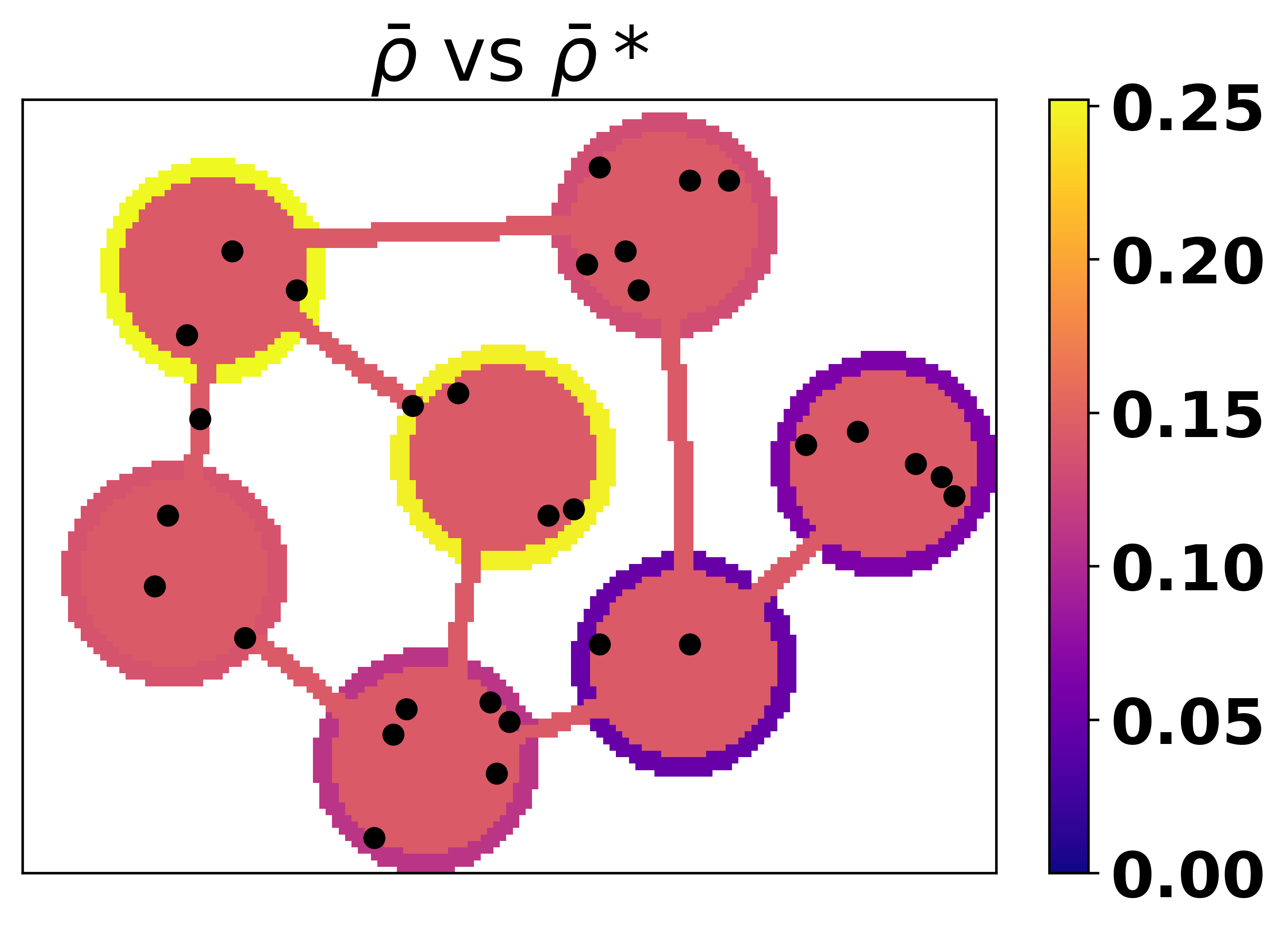}
            \caption{\(\beta = 0\)}
        \end{subfigure}
        ~
        \begin{subfigure}[t]{0.22\textwidth}
            \centering
            \includegraphics[width=1\textwidth]{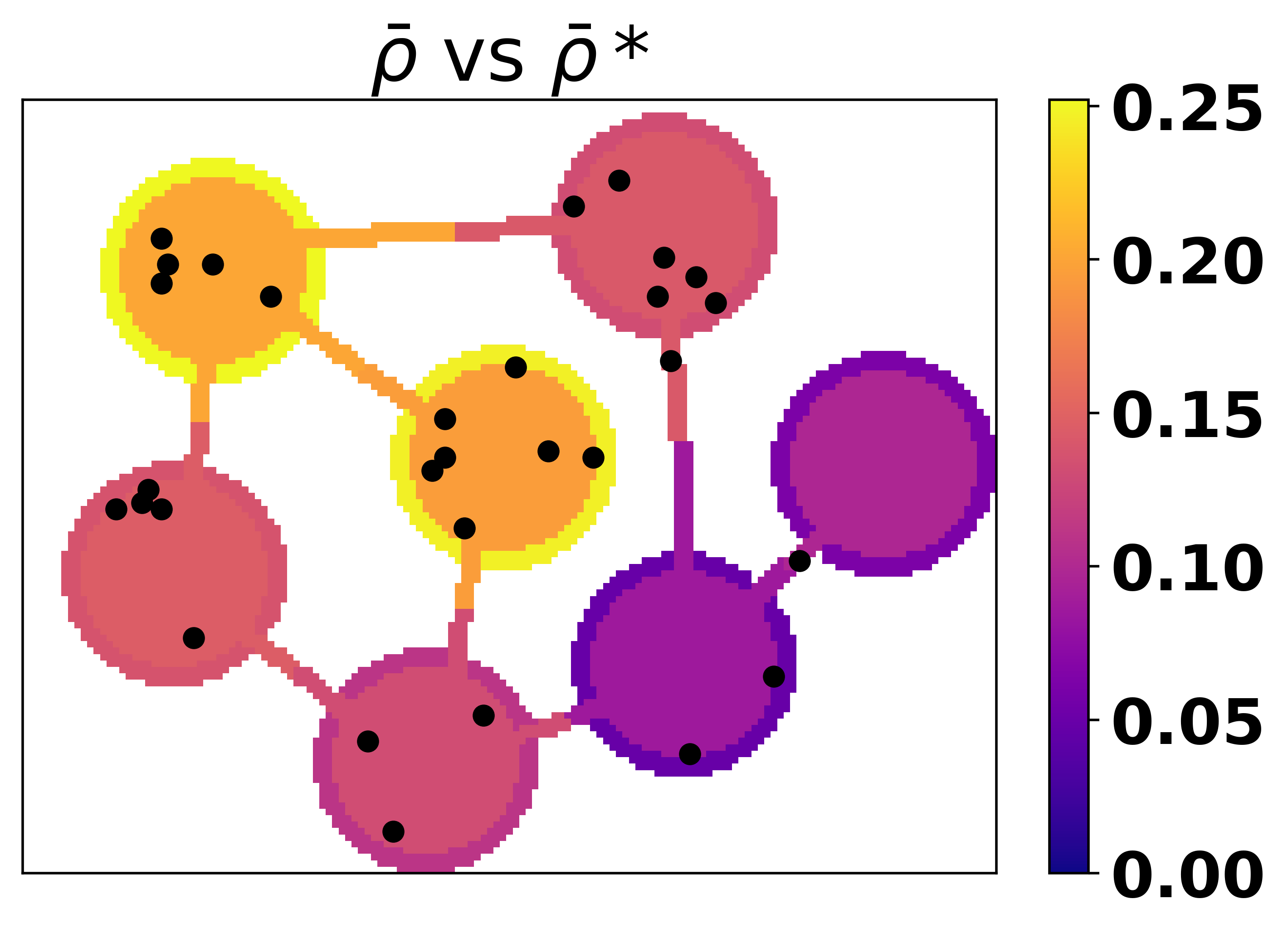}
            \caption{\(\beta = 0.5\)}
        \end{subfigure}%
        ~ 
        \begin{subfigure}[t]{0.22\textwidth}
            \centering
            \includegraphics[width=1\textwidth]{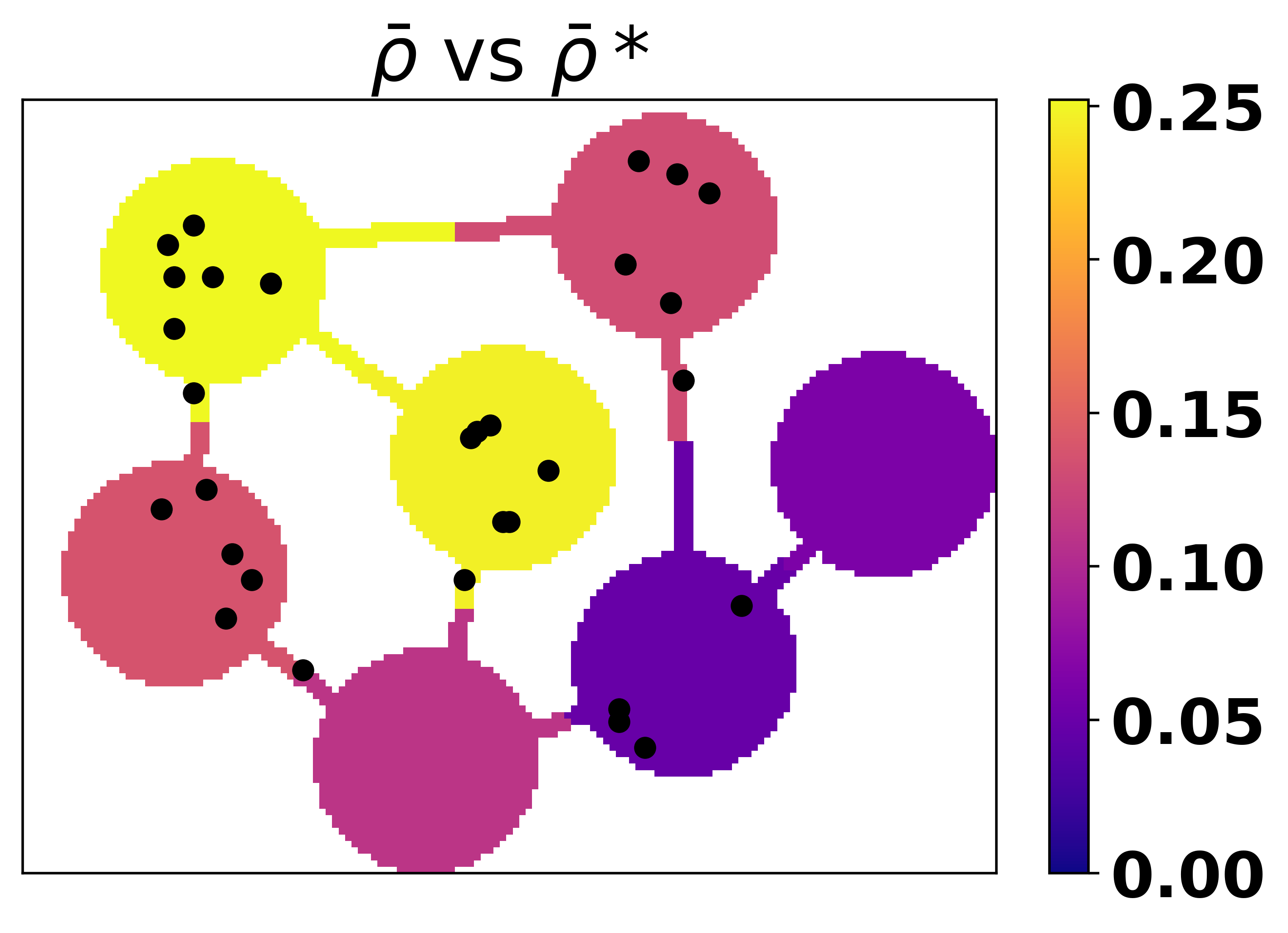}
            \caption{\(\beta = 1\)}
        \end{subfigure}
        ~ 
        \begin{subfigure}[t]{0.22\textwidth}
            \centering
            \includegraphics[width=1\textwidth]{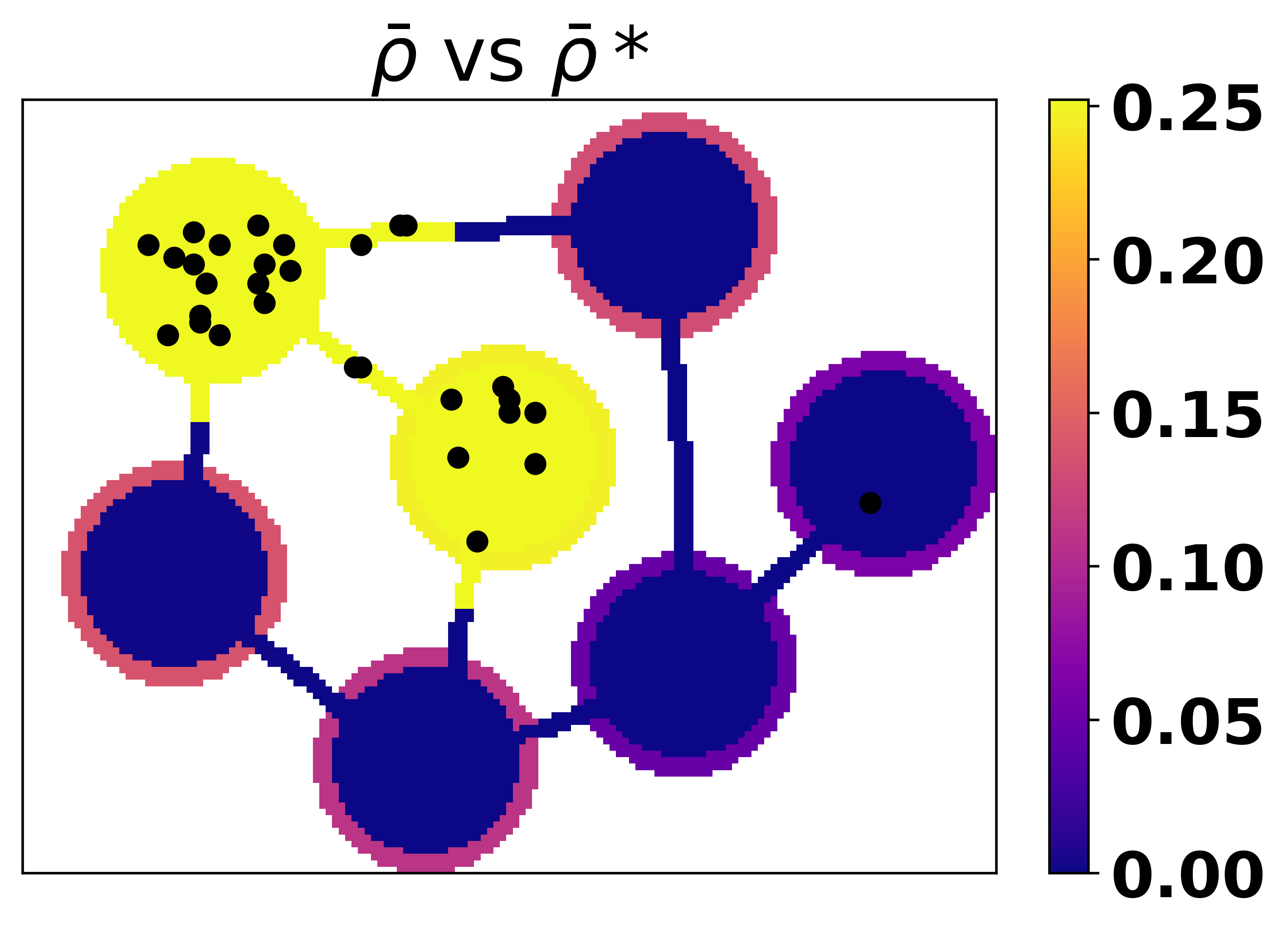}
            \caption{\(\beta = 10\)}
        \end{subfigure}
        \caption{Example distribution of 30 robots (black markers) with various coldness \(\beta\). The color of the border of the region represents the relative variance \(\sigma^2\), the inner color represents the target distribution \(\bar\rho(\beta)\). (a) the robots are equally spread out regardless of the variance; (b) the robots are somewhat between the uniform distribution and the optimal distribution; (c) the distribution of robots are proportional to the variance, which is the optimal for information gathering; (d) the robots are concentrated at the two regions with the highest variance, this causes severe under-sampling in the rest of the regions.} 
        \label{fig:coldness_example}
    \end{figure*}
    To account for the second subproblem of unreliable initial variance, the target distribution is designed to focus on collecting information on the variance at the beginning and gradually shift toward the oracle solution as the variance becomes reliable. This is achieved by simulated annealing, where the robots starts at the most random configuration, i.e. uniform random, and shift toward the optimal solution by varying a temperature parameter. 
    
    To achieve this, the \textit{Gibbs measure}\footnote{Also known as \textit{Boltzmann distribution in finite space}} is chosen as the measure \(\mu\) in \eqref{eq:graph_ergodicity}, with 
    \begin{equation}
    \begin{aligned}
    \mu(r_i) &= \exp(-\beta E(r_i))\\
    \end{aligned}
    \end{equation}
    where \(E(r_i)\) is the energy associated with the state \(r_i\), and \(\beta\) is the \textit{coldness} (or \textit{inverse temperature}) parameter. Then, since the region set is disjointed, the normalizing factor is 
    \begin{equation}
    \begin{aligned}
    \mu(\mathcal{R}) = \sum_{i=1}^n \exp(-\beta E(r_i)).
    \end{aligned}
    \end{equation}
    
    In particular, we define the energy to be the negative (differential) entropy of the sample distribution as
    \begin{equation}
    \begin{aligned}
    E(r_i) = -\ln(2\pi e\bar\sigma_i^2).
    \end{aligned}
    \end{equation}
    The space average in \eqref{eq:space_time_average} is then 
    \begin{equation}
    \begin{aligned}
    \bar\rho(\beta) &= \frac{1}{Z(\beta)}\sum_{i=1}^nI(r_i)\exp(\beta \ln(\bar\sigma_i^2))\\
    Z(\beta)  &= \sum_i \exp(\beta  \ln(\bar\sigma_i^2)).
    \end{aligned}
    \end{equation}

    An additional advantage of using the Gibbs measure is that the target distribution \(\bar\rho\) is guaranteed to be reachable by an ergodic Markov chain, as all the entries are strictly positive for any real-valued energy level \(E(r_i)\). More importantly, the coldness parameter \(\beta\) provides a smooth control on the uniformness of the target distribution, with \(\beta = 0 \) generating a uniform distribution regardless of the energy level; and \(\beta \rightarrow  \infty\) generating a delta function at the minimum energy state. 
    
    In our setup, the target distribution will assign more samples to regions with higher sampling entropy as \(\beta\) increases. Specifically, when \(\beta = 1\), the target distribution is 
    \begin{equation}
    \begin{aligned}
    \bar\rho(1) &= \frac{1}{Z(1)}\sum_{i=1}^nI(r_i)\exp(\ln(\bar\sigma_i^2))\\
    &=\frac{1}{Z(1)}\sum_{i=1}^nI(r_i)\bar\sigma_i^2.
    \end{aligned}
    \end{equation}
    If \(\bar\sigma^2 = \sigma^2\), then this is the estimation of the optimal sample size in \eqref{eq:optimal_dist}. An example of the effect of the value of \(\beta\) is shown in \figref{fig:coldness_example}. 
    As a result, if \(\beta\) varies gradually from \(0\) to \(1\), i.e., annealed, the goal of transitioning from uniform to optimal is achieved. In this work, annealing is done by the first-order step response as\footnote{Different annealing schedule can be used, such as tanh, with similar result. Here first-order step response is chosen with its similarity to Newton's law of cooling\cite[eq.(1.22)]{JohnHLienhard2013AHTT}.}
    \begin{equation}
    \begin{aligned}
    \beta(k) = 1 - \exp(-\alpha k),
    \end{aligned}
    \end{equation}
    with \(k\) being the time step and \(\alpha\) the cooling rate. The choice of 
    \(\alpha\) is discussed in Section \ref{section:cooling_rate}.

\section{Annealed Ergodic 
Algorithm }
   
    The complete algorithm for applying the annealed ergodic information gathering is shown in Algorithm \ref{alg:AEIG}.
    Since ergodic control can  run persistently, no time horizon has to be specified, and the only parameter required is the cooling rate \(\alpha\). The main algorithm is composed of two stages: a) sampling (lines 4 - 10), b) planning (lines 11 - 19). 
    Using the equations in \eqref{eq:mean_estimate} and \eqref{eq:variance_estimate} to estimate the mean and variance requires the entire set of observations to be stored. Instead, it is updated sequentially in a Bayesian manner using the normal-inverse gamma parameterization \((\nu, \mu, b)\) \cite[Section 6]{Murphy_2007}, where \(\mu\) is the mean, \(\nu\) is the number of samples, and \(b\) is an auxiliary variable to recover the variance, which is done using the equation in line 12. This framework can also be used to assign prior knowledge for the mean and variance. We use an uninformed prior and except for a well-defined initial variance, the parameter \(\nu\) is initialized at \(1\). Once the most recent estimated variance \(\sigma^2\) is recovered from the parameters, the target distribution \(\bar\rho\) is calculated using the current temperature \(\beta[k]\). The transition matrix \(P\) is then obtained using the REMC algorithm. The robots randomly transition to the next region individually according to the transition matrix. 
    
    \begin{algorithm}[!thpb]
    \caption{Annealed Ergodic Information Gathering }\label{alg:AEIG}
    \begin{algorithmic}[1]
    \State \textbf{Parameter:} Cooling Rate: \(\alpha\)
    \State \textbf{Initialize} \(\nu\gets \text{ones}(n), b\gets \text{ones}(n) \),\par\PH{Initialize}  \(\mu\gets \text{zeros}(n), k=0\)
    \While {inspecting}
    \For {each robot \(a\)}
        \State \(z = \text{Sample}(r(a))\)  \Comment{Sample from region of \(a\)}
        \State  \(\triangleright\) Update law for normal inverse-gamma dist.
        \State \(b[r(a)] \gets b[r(a)] + \frac{1}{2}\frac{n[r(a)]}{n[r(a)]+1}(z - \mu[r(a)])\) 
        \State \(\mu[r(a)] \gets \frac{n[r(a)]\mu[r(a)] + z}{n[r(a)]+1}\)
        \State \(\nu[r(a)] \gets \nu[r(a)]+1\) 
    \EndFor
    \For  { \(r_i \in \mathcal{R}\)}
        \State \(\sigma^2[i] \gets 2b[i]\frac{\nu[i]+1}{(\nu[i])^2}\)
    \EndFor
    \State \(\beta = 1 - \exp(-\alpha k)\)
    \State \(\bar\rho \gets \exp(\beta\ln(\sigma^2))\) \Comment{Entry-wise exponential}
    \State \(\bar\rho \gets \bar\rho/\sum_i \bar\rho_i\)
    \State \(P \gets \text{REMC}(\bar{\rho})\) 
    \For {each robot \(a\)}
    \State \(r[a] \gets P(r[a])\)
    \EndFor 
    \State \(k \gets k+1\)
    
    \EndWhile
    \end{algorithmic}
    \end{algorithm}

\section{Experiments}
    In the section, the annealing algorithm is compared against uniform coverage
     \begin{equation}
    \begin{aligned} 
    \bar\rho_\text{uniform} = \left[1/n ,1/n,  \cdots \right]
    \end{aligned}
    \end{equation}
    and direct ergodic coverage
         \begin{equation}
    \begin{aligned} 
    \bar\rho_\text{direct}(k) = \frac{1}{Z(1)}\sum_{i=1}^nI(r_i)\bar\sigma_i^2[k]
    \end{aligned}
    \end{equation}
    with different numbers of robots (5, 30) and ergodic Markov chain planning methods (REMC and FMMC). 

\subsection{REMC}
\subsubsection{Methodology}
    The simulation is performed on a graph structure identical to the graph in \figref{fig:example_task}. The observation noises \(\sigma^2\) are uniformly randomly generated from the range \((0, 20]\) and the mean are uniformly randomly generated from the range \((-10, 10]\). Both are sampled once at the beginning and used throughout the trials. Additionally, the noise is multiplied by the number of robots \(N\), this is emulating for a fix cost, one can get a large team of robot with bad sensors or a small team of robot with good sensors. All the robots start from the same region (region 1) to emulate the condition that the team is activated 
    at the same time from a base station. The annealing rate \(\alpha\) is chosen as \(0.025\). 100 trials are conducted for each method. The robots are modeled as collision-free point masses that synchronously transition to their planned regions at each time step \(k\). Parameter estimation (of \(\mu, \nu, b\)) and Markov chain optimization are performed in a centralized manner. The resultant stochastic matrix \(P\) is published to the robots and stochastic planning is carried out by the robots independently.  

\subsubsection{Results}
\begin{figure}[thpb!]
        \centering
        \includegraphics[width=0.45\textwidth]{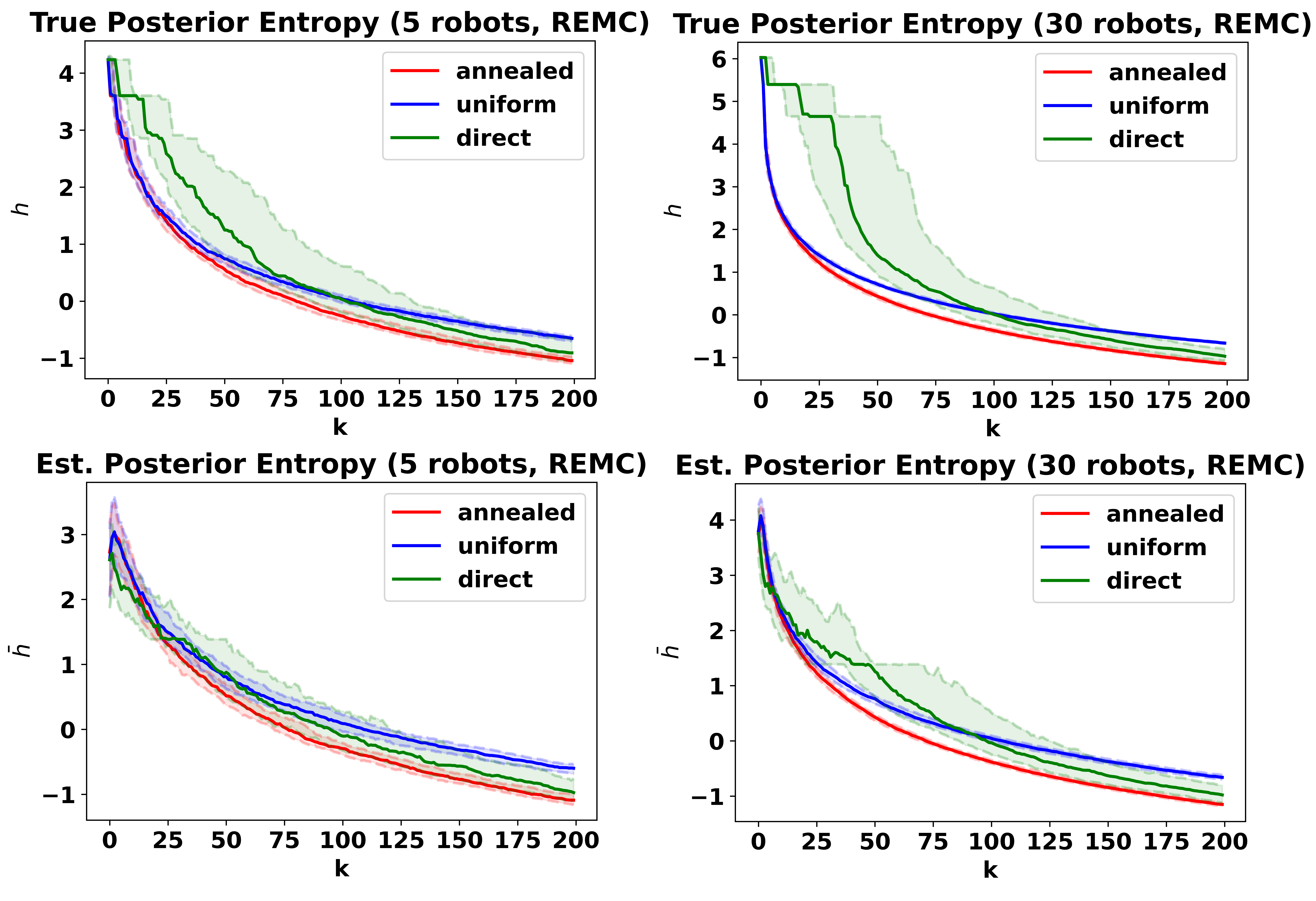}
        \caption{Comparison of maximum posterior entropy between uniform, direct ergodic, and annealed ergodic over 100 trials. With the solid line representing the median and the shaded region bounded by the dash line representing the first to third quartile region. top row shows the true entropy obtained from external oracle. This shows that annealing is consistently performing better both transiently and  asymptotically. bottom row shows the estimated entropy from the internal believe of the robots. This shows that when compared to the true entropy, direct ergodic method has a problem of overestimating the information it has. 
        }
        \label{fig:entropy_REMC}
    \end{figure}

    \figref{fig:entropy_REMC} shows that the annealing method outperforms the direct method during the transient phase, and the uniform method asymptotically with respect to the true posterior entropy. The true posterior entropy is obtained from the true sample variance \(\sigma^2\) and is maximized over the region, i.e. the region with the worst entropy is chosen for each time step \(k\), where 
    \begin{equation}
    \begin{aligned} 
    h(k) = \max_i \left[\ln\left(\frac{\sigma_i^2}{\nu_i(k)}\right)\right].
    \end{aligned}
    \end{equation}
    It is seen that the direct method has a true posterior entropy higher than both uniform and annealing methods from \(k=0\) to \(k=200\). This is caused by the previously mentioned problem of variance estimation error, which leads the robots to initially misallocate the samples until the variance is more refined. 
    Consequently, the direct method also has more variability over the trials (shown in the shaded region) because of its high dependence on the quality of initial observations. Conversely, while the uniform method shows a better initial performance, its long-term performance is plateaued by the regions with high observation noise, since all regions are always assigned the same number of samples with the maximum entropy dominated by the noisiest region. The annealing method shows advantages over both the uniform and direct methods. Initially, when the coldness parameter \(\beta = 0\), it closely follows the performance of the uniform method. As more samples are collected and \(\beta \rightarrow 1\), the robots shift toward the optimal distribution. 
    
    More insight can be gained from the bottom row of \figref{fig:entropy_REMC}, which plots the posterior entropy estimated by the robots. Here, the oracle \(\sigma^2\) is replaced by the estimated variance at each time step \(\bar\sigma^2(k)\) to yield
    \begin{equation}
    \begin{aligned} 
    \bar{h}(k) = \max_i\left[\ln\left(\frac{\bar\sigma_i^2(k)}{\nu_i(k)}\right)\right].
    \end{aligned}
    \end{equation}
    Uniform and annealed methods show similar estimated posterior entropy to the true posterior entropy, except they are nosier at the beginning when the sample size is small. However, direct entropy shows a significantly smaller estimated entropy than the true entropy. In other words, with the direct ergodic method, the robots believe they have more information than they really have, which directly causes the aforementioned misallocation of resources. Therefore, this shows that starting with uniform search is more advantageous as it avoids the problem of overconfidence.  
    
\begin{figure}[thpb]
      \centering
      \includegraphics[width=0.48\textwidth]{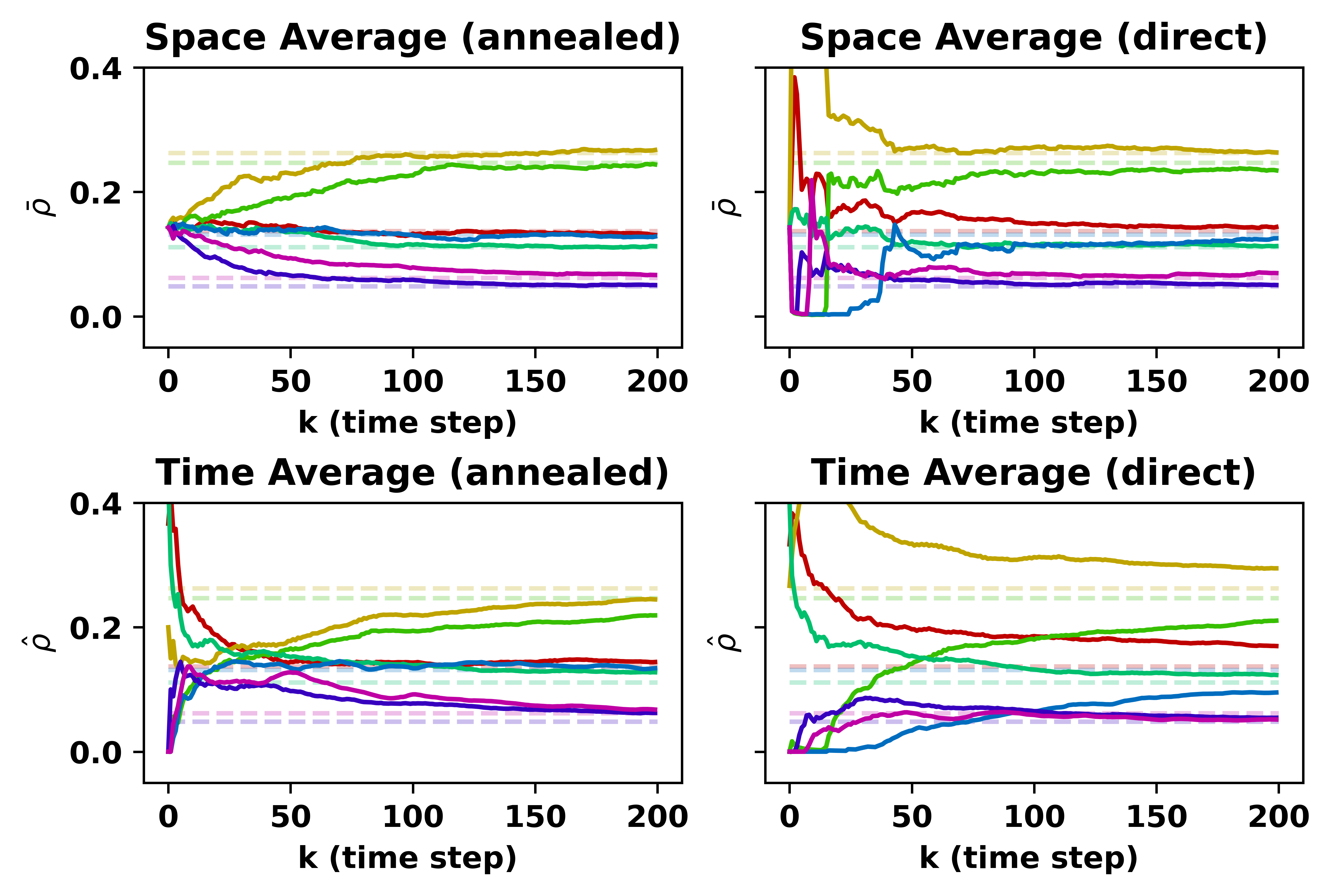}

      \caption{ Example space average \(\bar\rho\) and time average \(\hat\rho\) of each region shown in different color, with the optimal distribution shown in dashed-line. The left column shows the result using annealing and right column shows the result of direct ergodic. Annealing shows a smooth transition from uniform to the optimal solution which is trackable for the time average; while the direct ergodic method produced a fluctuating space average that causes an extreme time average.}
      \label{fig:example_space_vs_time}
   \end{figure}

    The region-wise behavior of the robots distribution can be seen in \figref{fig:example_space_vs_time}, which shows the space average and time average of one example trial from the annealed and direct methods. It can be seen that the annealed method generates smooth trajectories from uniform distribution to the optimal distribution for the time average to follow, while the direct method has a fluctuating space average caused by the initial noisy variance estimation with a sudden shift to the optimal distribution as the sample size increases. This causes a highly time-varying distribution that is difficult for the ergodic control to track and leads to the extreme values seen in the time average. 
   
\subsection{FMMC and Metropolis-Hasting}
    In this section, to explore the benefit of annealing to other graph traversal methods, the simulation 
    is run with the ergodic graph planner switched from REMC to other Markov chain algorithms capable of producing ergodic traversal. Specifically, FMMC \cite[Section 6]{alma99124373890001452}, which is optimized for convergence of \(\rho[k]\), instead of \(\hat\rho\), toward \(\bar\rho\); and Metropolis-Hasting\cite[Section 1.2.2]{alma99124373890001452}, which has the benefit of being ergodic without knowing the graph structure. Both the methods require a \textit{reversible Markov chain}, which has a negative impact on the convergence rate. 
    
    \subsubsection{Results}
\begin{figure}[thpb!]
        \centering
        \includegraphics[width=0.45\textwidth]{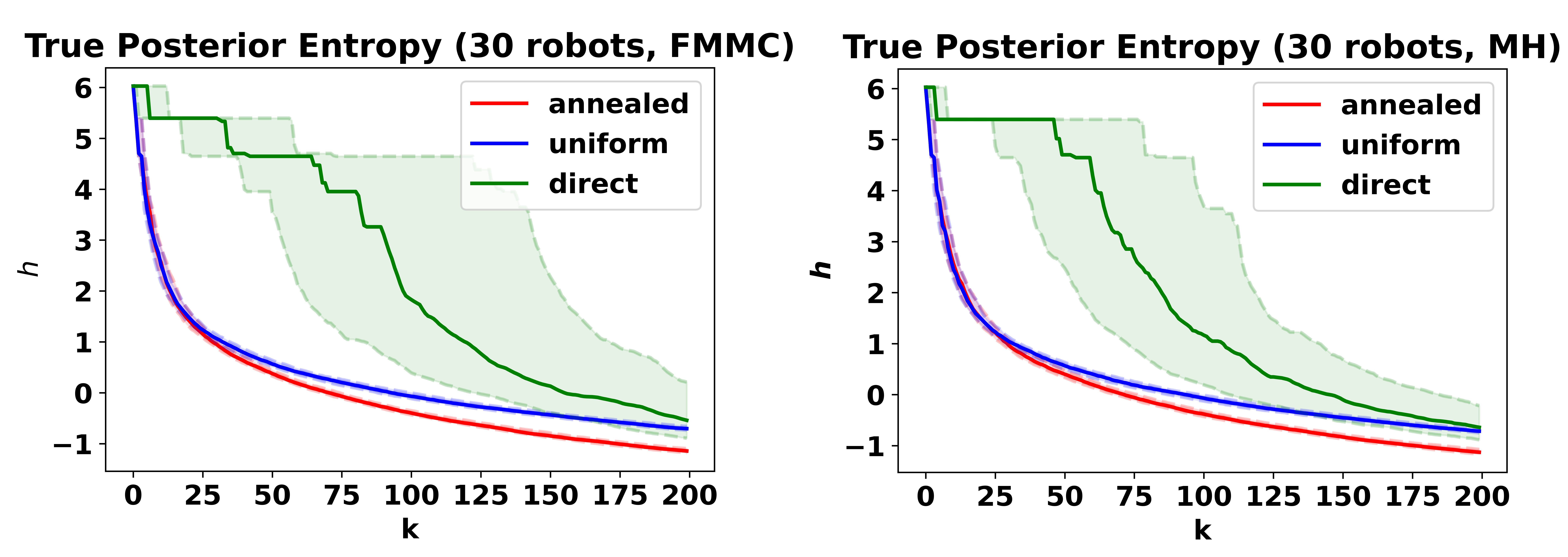}
        \caption{Posterior entropies similar to \figref{fig:entropy_REMC} with REMC ergodic planner replaced with FMMC on the left and
        Metropolis-Hasting (M-H) on the right. This shows more significant advantage of using annealing method under slower time-average convergence rate when compared to REMC.}
        \label{fig:entropy_FMMC}
    \end{figure}

    \figref{fig:entropy_FMMC} shows similar result with REMC in the previous section in the true posterior entropy. Since uniform is a static distribution, the convergence rate has less impact on performance. Similarly, the annealed method provides a smoothly varying target distribution that is easy for the Markov chain to track. Conversely, the direct method shows even worse performance compared to the previous section, with high variance across the trials, and an overall worse performance where the median performance fails to reach the performance of the annealed solution. This shows that the concept of annealed ergodic coverage provides an advantage to a wide spectrum of ergodic methods. 

\subsection{Physical Demonstration}
    \begin{figure*}[thpb!]
        \centering
         \begin{subfigure}[t]{0.22\textwidth}
            \centering
            \includegraphics[width=1\textwidth]{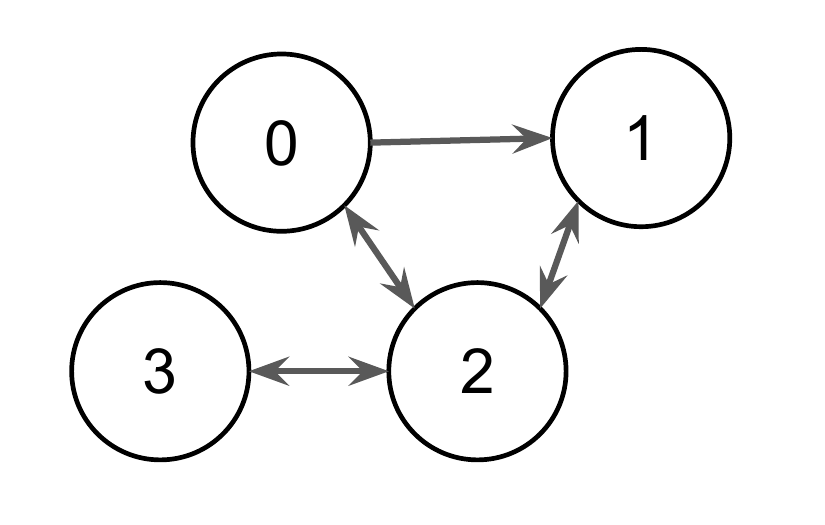}
            \caption{Graph structure}
        \end{subfigure}
        ~
        \begin{subfigure}[t]{0.22\textwidth}
            \centering
            \includegraphics[width=1\textwidth]{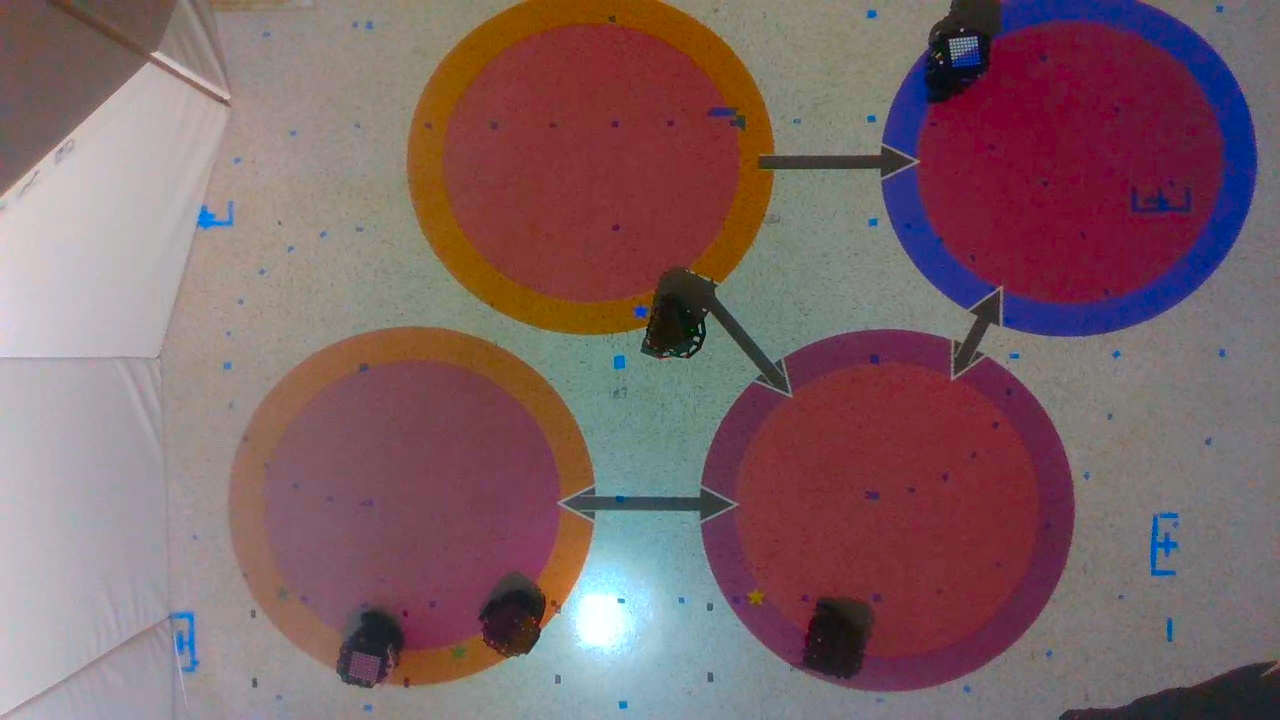}
            \caption{\(k = 3 \quad (\beta =0.26 )\)}
        \end{subfigure}%
        ~ 
        \begin{subfigure}[t]{0.22\textwidth}
            \centering
            \includegraphics[width=1\textwidth]{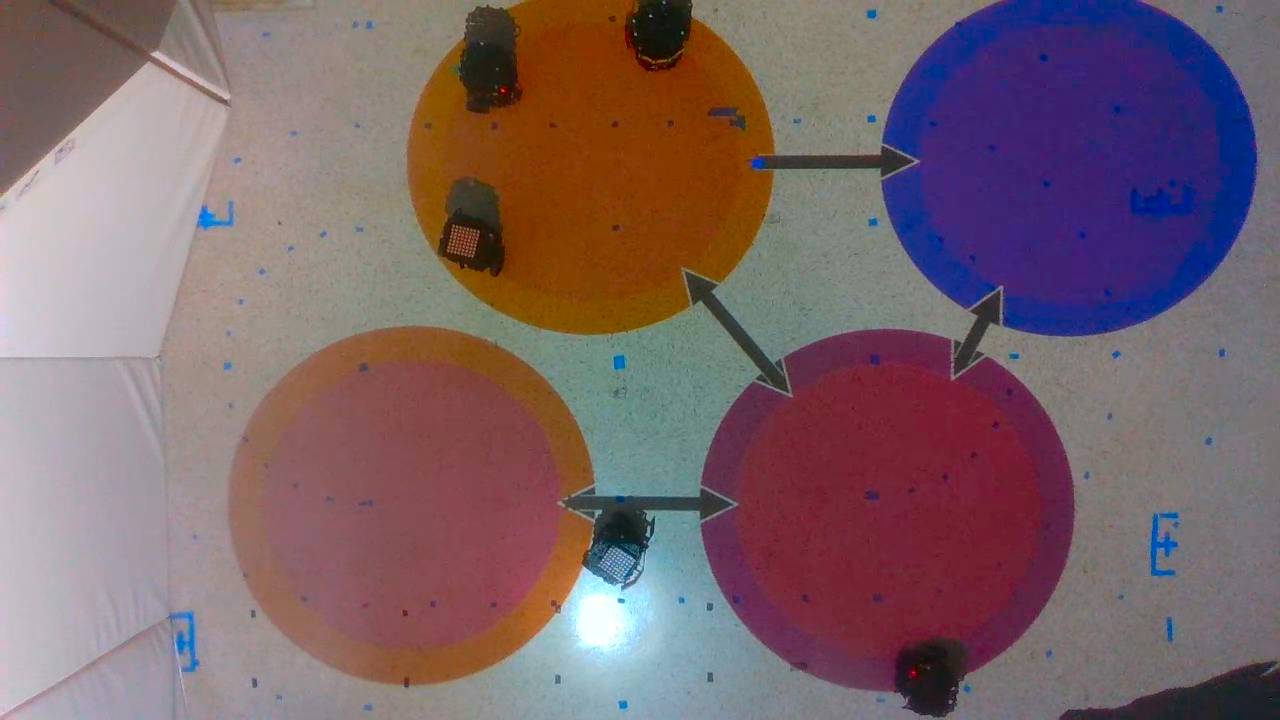}
            \caption{\(k = 13 \quad (\beta = 0.73)\)}
        \end{subfigure}
        ~ 
        \begin{subfigure}[t]{0.22\textwidth}
            \centering
            \includegraphics[width=1\textwidth]{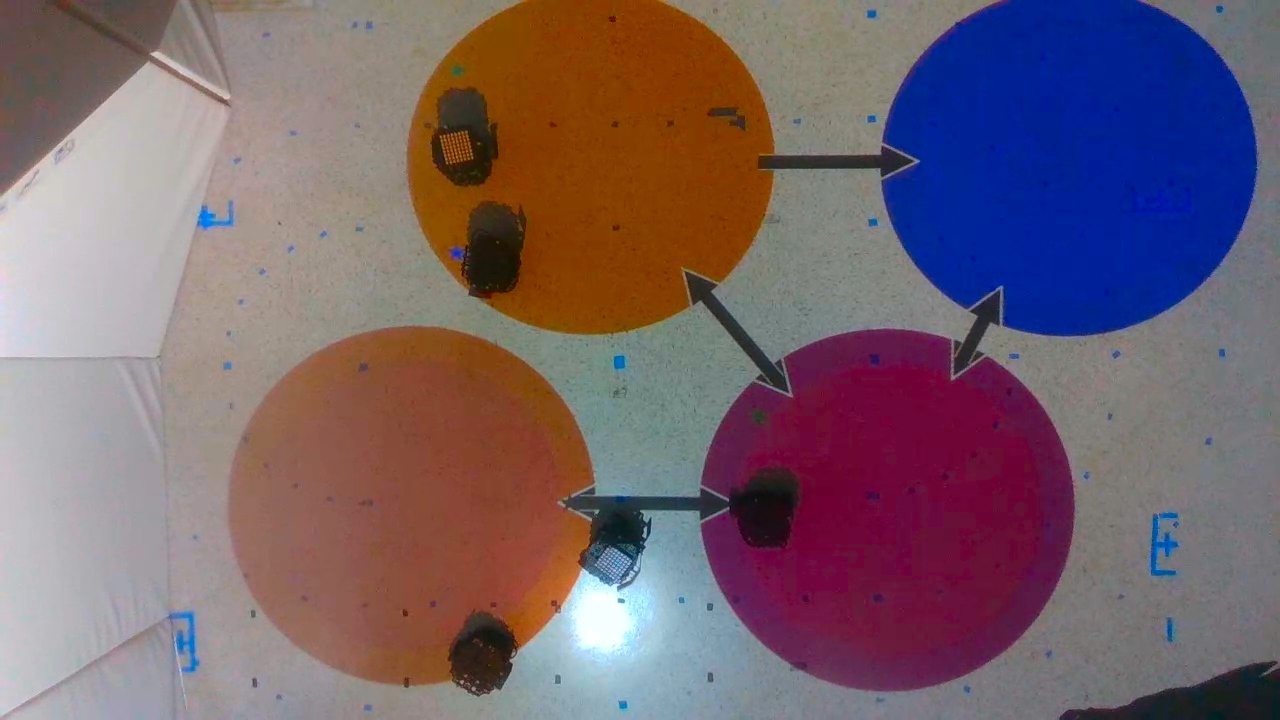}
            \caption{\(k = 29 \quad (\beta=0.94)\)}
        \end{subfigure}
        \caption{
        Experimental validation. (a) shows the graph structure of the system used for physical demonstration. (b)-(d) shows the distribution of the TurtleBot Burger robots at different representative timestamps during the annealing process. The fill colors in the circular regions match the border colors as \(\beta\) increases, showing that the system has correctly identified the optimal target distribution. } 
        \label{fig:drone_example}
    \end{figure*}
The simulation result is applied to a swarm system with an OptiTrack motion capture rig to validate the feasibility of executing the exploration on physical platforms. A total of five TurtleBot Burger robots are used with twelve Flex13 tracking cameras on the motion capture system. A 4-region graph is projected to the ground. The cooling rate \(\alpha\) is adjusted to \(0.1\). Snapshots of the system at 3 different time steps are shown in \figref{fig:drone_example}. The result shows that the planning pipeline can be executed in a physical platform with a suitable low-level collision avoidance technique.

\section{Discussion}
\label{section:cooling_rate}
An open question for the annealing process is the optimality of the cooling rate \(\alpha\). In this work, we choose \(\alpha\) empirically according to the simulation time horizon. Conceptually, the optimality of the cooling rate is correlated to the convergence rate of the ergodic planner, e.g. the ergodic controller should be able to reach uniform distribution before the annealing completely cools down. It is also correlated to the convergence of the noise estimation. Some observation models may require more samples to reach accurate noise estimation. The cooling rate then has to be slower so that the robots spend more time in the information collection phase. 

In this work, the target distribution is time-varying with the application of annealing and the updating variance \(\sigma^2\). This is not included in the proof for the optimality of REMC. Conceptually, since the variance estimation in \eqref{eq:variance_estimate} converges as \(\nu\) converges, and the annealing converges to the estimated variance, a convergence proof can be established. This is related to the notion of \textit{weak ergodicity}, which is defined for a time-varying Markov chain and is used to develop the convergence proof in simulated annealing \cite{BremaudPierre2013MCGF}. 

One main challenge of the practicality of this framework is the assumption that all agents can communicate with the command/control center at all time steps. The Markov chain-based ergodic planner can be used in a fully decentralized manner without any communication for a static target distribution. However, due to the information gathering task, the target distribution is constantly being updated, which requires a new Markov chain to be generated and broadcast to the robots. In our future work, we will investigate a fully decentralized framework using percolation theory \cite{Kim}. With the controller guaranteeing (weak) ergodicity, percolation theory can be applied \cite{Ghosh2025} and, ideally, a critical transition point can be established for information to be fully shared in the robot network without requiring a command center. 

For simplicity, the target information in this work is scalar and static. However, in many robotics applications, such as target tracking, the information will be multi-dimensional and dynamic. 
The core idea of ``lower the quality of information, more robots should be allocated'' should still be valid. Some extra considerations have to be accounted for, such as choosing a proper scalarization of variance (entropy, a-optimality, etc.) and establishing asymptotic confidence so that weak ergodicity holds true.

\section{Conclusion}
This paper addresses the problem of identifying a space average during multi-robot information acquisition such that ergodic control provides the theoretical optimal solution. It specifically formulates the acquisition problem on a region graph-based discretization of the environment, such that observation effort has to be allocated 
according to the quality of observed (sampled) information in the different regions. A multi-agent extension of a previously developed rapidly ergodic Markov chain planner is introduced to solve 
the problem. The problem of unknown quality of information is further addressed by introducing a simulated annealing-based space average. The estimated entropy is applied to the Boltzmann distribution such that annealing can be done by gradually varying the coldness parameter from 0 to 1, which controls the space average from a uniform distribution to the optimal distribution. Substantial performance improvements 
over uniform and direct ergodic search methods are shown for varying number of robots on an example graph in simulation. The performance benefits are also validated on a ground robot swarm system.


\bibliographystyle{IEEEtran} 
\bibliography{refs}
\clearpage

\end{document}